\documentclass[bst/sn-mathphys]{sn-jnl}

\jyear{2021}%

\theoremstyle{thmstyleone}%
\usepackage{graphicx}
\usepackage{amsmath}
\usepackage{booktabs}
\usepackage{amssymb}
\usepackage{multirow}
\usepackage{multicol}
\usepackage{makecell}
\usepackage{overpic}
\usepackage{bbding}
\usepackage{wasysym}
\usepackage{arydshln}
\usepackage{latexsym}

\theoremstyle{thmstyletwo}%

\theoremstyle{thmstylethree}%

\begin{document}

\title[VLP: A Survey on Vision-Language Pre-training]{VLP: A Survey on Vision-Language Pre-training}


\author[1,2]{\fnm{Feilong} \sur{Chen}}
\equalcont{These authors contributed equally to this work.}

\author[1,3]{\fnm{Duzhen} \sur{Zhang}}
\equalcont{These authors contributed equally to this work.}

\author[1,3]{\fnm{Minglun} \sur{Han}}
\equalcont{These authors contributed equally to this work.}

\author[1,3]{\fnm{Xiuyi} \sur{Chen}}

\author[1]{\fnm{Jing} \sur{Shi}}
\author[1]{\fnm{Shuang} \sur{Xu}}
\author[1,2,3]{\fnm{Bo} \sur{Xu}}

\affil[1]{\orgname{Institute of Automation, Chinese Academy of Sciences}, \orgaddress{\city{Beijing} \postcode{100190},  \country{China}}}

\affil[2]{\orgdiv{School of Future Technology}, \orgname{University of Chinese Academy of Sciences}, \orgaddress{\city{Beijing} \postcode{100049},  \country{China}}}

\affil[3]{\orgdiv{School of Artificial Intelligence}, \orgname{University of Chinese Academy of Sciences}, \orgaddress{\city{Beijing} \postcode{100049},  \country{China}}}


\abstract{In the past few years, the emergence of pre-training models has brought uni-modal fields such as computer vision (CV) and natural language processing (NLP) to a new era. Substantial works have shown they are beneficial for downstream uni-modal tasks and avoid training a new model from scratch. So can such pre-trained models be applied to multi-modal tasks? Researchers have explored this problem and made significant progress. This paper surveys recent advances and new frontiers in vision-language pre-training (VLP), including image-text and video-text pre-training. To give readers a better overall grasp of VLP, we first review its recent advances from five aspects: feature extraction, model architecture,  pre-training objectives, pre-training datasets, and downstream tasks. Then, we summarize the specific VLP models in detail. Finally, we discuss the new frontiers in VLP. To the best of our knowledge, this is the first survey focused on VLP. We hope that this survey can shed light on future research in the VLP field.}

\keywords{Vision and language, Pre-training, Transformers}



\maketitle

\section{Introduction}\label{sec1}
Making machines respond in ways similar to humans has been a relentless goal of AI researchers. To enable machines to perceive and think, researchers propose a series of related tasks, such as face recognition, reading comprehension, and human-machine dialogue, to train and evaluate the intelligence of machines in a particular aspect. Specifically, domain experts manually construct standard datasets and then train and evaluate relevant models on them. However, due to the limitations of related technologies, it is often necessary to train on a large amount of labelled data to obtain a better and more capable model. The recent emergence of pre-training models based on the Transformer structure \cite{vaswani2017attention} has alleviated this problem. They are first pre-trained via self-supervised learning that typically exploits auxiliary tasks (pre-training objectives) to mine supervision signals from large-scale unlabelled data to train the model, thereby learning universal representations. Then they can achieve surprising effectiveness by fine-tuning with only a tiny amount of manually-labelled data on downstream tasks. Since the advent of BERT \cite{DBLP:conf/naacl/DevlinCLT19} in natural language processing (NLP), various pre-training models have sprung up in the uni-modal field, such as Vision Transformer (ViT) \cite{dosovitskiy2020image} in computer vision (CV) and Wave2Vec \cite{DBLP:conf/interspeech/SchneiderBCA19} in speech. Substantial works have shown they are beneficial for downstream uni-modal tasks and avoid training a new model from scratch.

Similar to the uni-modal field, there is also a problem of less high-quality labelled data in the multi-modal field. The natural question is, can the above pre-training method be applied to multi-modal tasks? Researchers have explored this problem and made significant progress. In this paper, we focus on mainstream vision-language pre-training (VLP), including image-text and video-text pre-training. VLP mainly learns the semantic correspondence between different modalities by pre-training on large-scale data. For example, in image-text pre-training, we expect the model to associate ``dog'' in text with what ``dog'' looks like in images. In video-text pre-training, we expect the model to map objects/actions in the text to objects/actions in the video. To achieve this goal, the VLP objects and model architecture need to be cleverly designed to allow the model to mine the associations between different modalities.

To give readers a better global grasp of VLP, we first comprehensively review its recent advances and focus on five significant aspects:
\begin{itemize}
    \item \textbf{Feature extraction.} This section includes the preprocessing and representation methods of image, video, and text in VLP models (see Section \ref{fe}).
    \item \textbf{Model architecture.} We introduce the architecture of the VLP models from two different perspectives: Single-stream versus Dual-stream from multi-modal fusion perspective, and Encoder-only versus Encoder-decoder from the overall architectural design perspective (see Section \ref{ma}).
    \item \textbf{Pre-training objectives.} Pre-training objectives are the core of VLP, mainly used to guide the model to learn vision-language associated information. We summarize typical and characteristic pre-training objectives divided into completion, matching, temporal, and particular types (see Section \ref{po}).
    \item \textbf{Pre-training datasets.} Data is critical for VLP. We briefly introduce mainstream corpora for VLP and their specific sizes (see Section \ref{pd}).
    \item \textbf{Downstream tasks.} Various tasks requires a cooperative knowledge of both vision and language. We discuss the basic details and goals of these tasks (see Section \ref{dt}).
\end{itemize}
Then we summarize the specific state-of-the-art (SOTA) VLP models in detail (see Section \ref{vlp}). Finally, We conclude the paper and have broad discussions on new frontiers in VLP (see Section \ref{nf}).

Although there are many surveys on pretrained language models~\cite{qiu2020pre,han2021pre} and pretrained vision models~\cite{han2022survey}, to the best of our knowledge, this is the first survey focused on VLP. We hope that our survey can help researchers better understand this field and inspire them to design better models.

\section{Feature Extraction} \label{fe}
This section describes how VLP models preprocess and represent an image, video and text to obtain counterpart features.

\subsection{Feature Extraction}
\subsubsection{Image Feature Extraction}
\paragraph{(1) OD-based Region Features (OD-RFs).} Most previous work~\cite{lu2019vilbert,li2019visualbert,li2020oscar} on VLP utilizes pre-trained object detectors to extract visual features. The most commonly used object detection model is Faster R-CNN~\cite{ren2015faster} with bottom-up attention~\cite{anderson2018bottom}. It is designed to identify objects belonging to certain classes and localize them with bounding boxes. By using the Faster R-CNN, VLP models obtain the OD-based Region feature embedding $V = [o_1, o_2, \dots, o_k]$ of an image with $k$ selected regions. Each region feature $o_i$ is a $2048$-d Region-of-Interest (RoI) feature with its bounding box. The bounding box is defined by the coordinates of the bottom-left and top-right corners of the region. VLP models use bounding boxes to construct $5$-d vectors, and the vector is embedded into a high-dimensional representation (2048-d) named visual geometry embedding. The OD-RFs are obtained by adding the OD-based Region feature embedding with its visual geometry embedding. Although ODFs have brought impressive performance, extracting region features can be time-consuming. To relieve this problem, the pre-trained object detectors are usually frozen during pre-training, which can limit the capacity of VLP models.

\paragraph{(2) CNN-based Grid Features (CNN-GFs).} VLP models~\cite{li2020unicoder,wang2021simvlm} extract visual features by utilizing convolutional neural networks (CNNs) to obtain the grid features. On the one hand, VLP models can train the CNNs end-to-end by using the grid features~\cite{jiang2020defense} directly. On the other hand, VLP models can also first discretize grid features using a learned vision dictionary, then feed them into the cross-modal module.

\paragraph{(3) ViT-based Patch Features (ViT-PFs).} Inspired by ViT~\cite{dosovitskiy2020image,radford2021learning}, VLP models reshape the image $I_i \in \mathbb{R}^{H \times W \times C}$ into a sequence of flattened 2D patches ${I}_p \in \mathbb{R}^{N \times (P^2 \cdot C)}$, where $(H, W)$ is the resolution of the original image, $C$ is the number of channels, $(P,P)$ is the resolution of each image patch, and $N=HW/P^2$ is the resulting number of patches, which also serves as the effective input sequence length for the Transformer. An input image $I_i$ is encoded into a sequence of embeddings:
$\{{v}_\mathrm{cls}, {v}_1,...,{v}_N\}$, where $v_\mathrm{cls}$ is the embedding of the \texttt{[CLS]} token.

\subsubsection{Video Feature Extraction}
A video clip is denoted as $M$ frames (images). VLP models~\cite{luo2021clip4clip,fang2021clip2video} extract the frame features by using the method mentioned above. The two most commonly used features are CNN-GFs and ViT-PFs. For CNN-GFs, VLP models first use ResNet~\cite{he2016deep} pre-trained on ImageNet~\cite{deng2009imagenet} or SlowFast~\cite{feichtenhofer2019slowfast} and I3D~\cite{carreira2017quo} pre-trained on Kinetics~\cite{kay2017kinetics} to extract 2D and 3D visual features for each video frame. These features are concatenated as visual features and fed through a fully-connected (FC) layer to be projected into the same lower-dimensional space as token embeddings. For ViT-PFs, a video clip $V_i \in \mathbb{R}^{M \times H \times W \times C}$ consisting of $M$ frames of resolution $H \times W$, where $M = 1$ for images. Following the protocol in ViT and Timesformer, the input video clip is divided into $M \times N$ non-overlapping spatio-temporal patches of size $P \times P$, where $N = HW/P^2$.

\subsubsection{Text Feature Extraction}
For the textual features, following pretrained language model such as BERT~\cite{DBLP:conf/naacl/DevlinCLT19}, RoBERTa~\cite{liu2019roberta}, AlBERT~\cite{lan2019albert}, and XLNet~\cite{yang2019xlnet}, VLP models~\cite{li2019visualbert,zhang2021vinvl,zeng2021multi} first segment the input sentence into a sequence of subwords. And then, insert a start-of-sequence token and an end-of-sequence token at the beginning and the end of the sequence to generate the input text sequence. Text input representations are computed via summing the corresponding word embedding, text position embedding, and text type embedding.

\subsection{Feature Representation}
To make full use of uni-modal pre-trained models, VLP models can send the visual or text features to a transformer encoder~\cite{vaswani2017attention}. Specifically, VLP models utilize the standard transformer encoder with random initialization to generate the visual or textual representation. In addition, VLP models can utilize a pre-trained visual transformer to encode the ViT-PFs, such as ViT and DeiT~\cite{pmlr-v139-touvron21a}. VLP models can use a pre-trained textual transformer to encode the textual features, such as BERT. For simplicity, we name these transformer Xformer.

\section{Model Architecture}\label{ma}
In this section, we introduce the architecture of the VLP models from two different perspectives: (1) Single-stream versus Dual-stream from multi-modal fusion perspective, and (2) Encoder-only versus Encoder-decoder from the overall architectural design perspective.

\begin{figure}[t]
\centering
\scalebox{0.90}{
  \begin{overpic}[width=\textwidth]{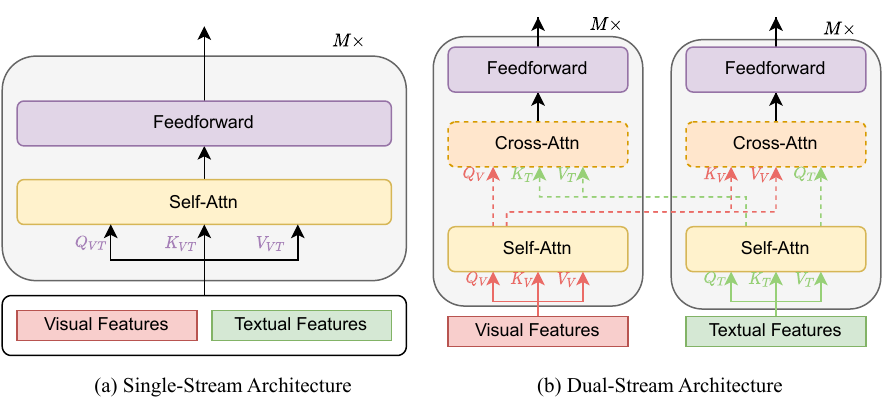}
  \end{overpic}
  }
  \caption{Illustration of two types of model architectures for VLP.
  }\label{fig:model_arch} 
\end{figure}

\subsection{Single-stream versus Dual-stream}
\paragraph{Single-stream Architecture.} The single-stream architecture~\cite{li2019visualbert,chen2020uniter,chen2022improving} refers to that the text and visual features are concatenated together, then fed into a single transformer block as shown in Firgue~\ref{fig:model_arch} (a). The single-stream structure utilizes merged attention to fuse multimodal inputs. The single-stream architecture is more parameter-efficient, as the same set of parameters is used for both modalities. 

\paragraph{Dual-stream Architecture.} The dual-stream architecture~\cite{zhang2020devlbert,dou2021empirical} refers to that the text and visual features are not concatenated together but sent to two different transformer blocks independently, as shown in Firgue~\ref{fig:model_arch} (b). These two transformer blocks do not share parameters. To achieve higher performance, cross-attention (as shown by the dotted line in Firgue~\ref{fig:model_arch} (b)) are used to enable cross-modal interaction. To achieve higher efficiency, there can also be no cross-attention between the visual transformer and textual transformer blocks.   

\subsection{Encoder-only versus Encoder-decoder}
Many VLP models adopt the encoder-only architecture, where the cross-modal representations are directly fed into an output layer to generate the final outputs. In contrast, other VLP models advocate using a transformer encoder-decoder architecture, where the cross-modal representations are first fed into a decoder and then to an output layer.

\section{Pre-training Objectives}\label{po}
This section introduces how we pre-train VLP models by using different pre-training objectives, which are crucial for learning the universal representation of vision-language. We summarize the pre-training objectives into four categories: completion, matching, temporal, and particular types.
\begin{itemize}
    \item \textbf{Completion} is to reconstruct the masked element by leverage the unmasked remainders to understand the modality. (see section~\ref{mlm},~\ref{plm} and ~\ref{mvm}).
    \item \textbf{Matching} is to unify the vision and language into a shared hidden space to generate universal vision-language representation (see Section~\ref{vlm},~\ref{vlc} and~\ref{wra}). 
    \item \textbf{Temporal} is to learn good representation by reorder the disrupted input sequence (see Section~\ref{foa})
    \item \textbf{Particular} types consists of other pre-training objects, such as visual question answering and visual captioning (see Section~\ref{vqa}).  
\end{itemize}
Now we introduce the most used pre-training objectives. 

\subsection{Masked Language Modeling}\label{mlm}
Masked language modeling (MLM), which was first proposed by Talylor~\cite{taylor1953cloze} in the literature, is widely known because the BERT model adapted it as a novel pre-training task. To model language conditioned on vision, MLM in VLP models is similar to MLM in pre-training language models (PLMs) but predicts the masked textual tokens not only by the rest of the textual tokens but also by the visual tokens. Empirically, VLP models following BERT randomly mask each textual input token with probability 15\% and replace the masked one by using a special token \texttt{[MASK]} 80\% of the time, a random textual token 10\% of the time and the original token 10\% of the time to perform masking. The formal definition is as follows:
\begin{equation}\label{eq:MLM}
  \mathcal{L}_{\rm MLM} = -{\rm E}_{(\mathbf{v}, \mathbf{w})\sim D}\log P(\mathbf{w}_m \vert \mathbf{w}_{\backslash m}, \mathbf{v}),
\end{equation}
where $\mathbf{v}$ denotes the vision, $\mathbf{w}$ denotes the textual tokens, $\mathbf{w}_m$ denotes the masked textual tokens, $\mathbf{w}_{\backslash m}$ denotes the remained textual tokens and $D$ denotes the training dataset.

\subsection{Prefix Language Modeling}\label{plm}
Prefix Language Modeling (PrefixLM)~\cite{wang2021simvlm} is unified of MLM and language modeling (LM). To make the model simultaneously has good understanding and generation ability, PrefixLM is proposed to facilitate the model with solid generation capability that enables text-induced zero-shot generalization without finetuning. PrefixLM differs from the standard LM such that it enables bi-directional attention on the prefix sequence and only conducts autoregressive factorization on the remaining tokens. PrefixLM under the sequence-to-sequence (seq2seq) framework not only enjoys the bidirectional contextualized representation as in MLM but also can perform text generation similar to LM. The formal definition is as follows:
\begin{equation}\label{eq:PrefixLM}
  \mathcal{L}_{\rm PrefixLM} = -{\rm E}_{(\mathbf{v}, \mathbf{w})\sim D}\log P(\mathbf{w}_{\ge T_p} \vert \mathbf{w}_{\le T_P}, \mathbf{v}),
\end{equation}
where $T_P$ denotes the length of the prefix sequence.

\subsection{Masked Vision Modeling}\label{mvm}
To have good understanding on vision or generate images/videos given text, like MLM, masked vision modeling (MVM)~\cite{chen2020uniter} samples vision (image or video) regions or patches and usually masks their visual features with a probability of 15\%. VLP models need to reconstruct the masked visual features given the remaining visual features and all the textual features. The masked visual features are set to zeros. Because visual features are high-dimensional and continuous, VLP models propose two variants for MVM.

\paragraph{(1) Masked Features Regression} learns to regress the model output of masked features to its original visual features. VLP models convert the model output of the masked features to a vector of the same dimension as the original visual features first and apply L2 regression between the original visual features and the vector. The formal definition is as follows:
\begin{equation}\label{eq:MVR1}
  \mathcal{L}_{\rm MVM} = {\rm E}_{(\mathbf{v}, \mathbf{w})\sim D} f(\mathbf{v}_m \vert \mathbf{v}_{\backslash m}, \mathbf{w}),
\end{equation}
\begin{equation}\label{eq:MFR}
  f(\mathbf{v}_m \vert \mathbf{v}_{\backslash m}, \mathbf{w}) = \sum_{i=1}^K \Vert h(\mathbf{v}_m^{i})-O(\mathbf{v}_m^{i}))\Vert_2^2,
\end{equation}
where $h(\mathbf{v}_m^{i})$ denotes the predicted vision representation and $O(\mathbf{v}_m^{i})$ denotes the original vision representation. 

\paragraph{(2) Masked Feature Classification} learns to predict the object semantic class for the masked features. VLP models first feed the output of the masked features into an FC layer to predict the scores of object class, which further goes through a softmax function to be transformed into a prediction normalized distribution. Note that there is no ground-truth label. There are two kinds of methods to train VLP models. One is that VLP models take the most likely object class from the object detection model as the hard label (w.p. 0 or 1), assuming the detected object class is the ground-truth label for the masked features and apply cross-entropy loss to minimize the gap between the prediction and pseudo class. The other is that VLP models utilize soft label as supervision signal, which is the raw output from the detector (i.e., a distribution of object classes) and minimize the KL divergence between two distributions.  The formal definition is as follows:
\begin{equation}\label{eq:MVR2}
  \mathcal{L}_{\rm MVM} = {\rm E}_{(\mathbf{v}, \mathbf{w})\sim D} f(\mathbf{v}_m \vert \mathbf{v}_{\backslash m}, \mathbf{w}).
\end{equation}
We use the object detection output from Faster R-CNN, and take the detected object category as the label of the masked region:

\begin{equation}\label{eq:MFR}
  f_1(\mathbf{v}_m \vert \mathbf{v}_{\backslash m}, \mathbf{w}) = \sum_{i=1}^K{\rm CE}(c(\mathbf{v}_m^{i})-g_1(\mathbf{v}_m^{i}))),
\end{equation}
where $g_1(\mathbf{v}_m^{i})$ the detected detected object category and $K$ denotes the number of vision regions.

We avoid this assumption by using soft label as supervision signal, which is the raw output from the detector:
\begin{equation}\label{eq:MFR}
  f_2(\mathbf{v}_m \vert \mathbf{v}_{\backslash m}, \mathbf{w}) = \sum_{i=1}^K{\rm D}_{KL}(\hat{c}(\mathbf{v}_m^{i})-g_2(\mathbf{v}_m^{i}))).
\end{equation}
where $g_1(\mathbf{v}_m^{i})$ the detected detected object category distribution.

\subsection{Vision-Language Matching}\label{vlm}
Vision-Language Matching (VLM)~\cite{li2021align} is the most commonly used pre-training objective to align vision and language, which aims to project vision and language into the same space. In the single-stream VLP models, they use the representation of the special token \texttt{[CLS]} as the fused representation of both modalities. In the dual-stream VLP models, they concatenate the visual representation of the special visual token \texttt{[CLS$_V$]} and the textual representation of the special textual token \texttt{[CLS$_T$]} as the fused representation of both modalities. VLP models feed the fused representation of both modalities to an FC layer and a sigmoid function to predict a score between 0 and 1, where 0 indicates the vision and language are mismatched, and 1 indicates the vision and language are matched. During training, 
VLP models sample positive or negative pairs from the dataset at each step. The negative pair is created by replacing the vision or text in a paired sample with randomly selected from other samples.

\subsection{Vision-Language Contrastive Learning}\label{vlc}
Vision-Language Contrastive Learning (VLC)~\cite{li2021align} also aims to align vision and language. Different VLM, VLC predicts the matched vision-language pairs from $N \times N$ possible vision-language pairs given a batch of $N$ vision-language pairs. Note that there are $N^2-N$ negative vision-language pairs within a training batch. VLP models use the visual representation of the special visual token \texttt{[CLS$_V$]} and the textual representation of the special textual token \texttt{[CLS$_T$]} to denote the aggregated representation of the vision and language, respectively. VLP models compute the softmax-normalized vision (image or video)-to-text similarity and text-to-vision similarity and leverage cross-entropy losses over vision-to-text and text-to-vision similarities to update themselves. The similarity is often implemented by dot products. The formal definitions are as follows:

\begin{equation}\label{eq:MFR}
  p^{v2t}_m(I) = \frac{{\rm exp}(s(I,T_m)/\tau)}{\sum_{m=1}^M{\rm exp}(s(I,T_m)/\tau)},
\end{equation}

\begin{equation}\label{eq:MFR}
  p^{t2v}_m(T) = \frac{{\rm exp}(s(T,I_m)/\tau)}{\sum_{m=1}^M{\rm exp}(s(T,I_m)/\tau)},
\end{equation}

\begin{equation}\label{eq:MFR}
  \mathcal{L}_{\rm VLC} = \frac{1}{2}{\rm E}_{(I,T)\sim D}[{\rm CE}(y^{v2t},p^{v2t}(I))+{\rm CE}(y^{t2v},p^{t2v}(T)],
\end{equation}
where $I$. $T$ denotes the images and texts, $s(\cot)$ denotes the similarity function and $\tau$ denotes temperature coefficient. $y^{v2t}$ and $y^{t2v}$ denote the labels of vision2text retrieval and text2vision retrieval.

\subsection{Word-Region Alignment}\label{wra}
Word-Region Alignment (WRA)~\cite{chen2020uniter} is an unsupervised pre-training objective to align vision regions (vision patches) and words. VLP models utilize Optimal Transport to learn the alignment between vision and language. Empirically, VLP models use the IPOT algorithm to approximate the OT distance since the exact minimization is computationally intractable. After solving minimization, the OT distance serves as the WRA loss to train VLP models. The formal definition is as follows:
\begin{equation}\label{eq:MVR2}
  \mathcal{L}_{\rm WRA} = \underset{{\rm T}\in \uppercase\expandafter{\romannumeral2}(\mathbf{a},\mathbf{b})}{\rm min}\sum_{i=1}^T\sum_{j=1}^K {\rm T}_{ij}\cdot c(\mathbf{w}_i, \mathbf{v}_j),
\end{equation}
where $c(\mathbf{w}_i, \mathbf{v}_j)$ is the cost function evaluating the distance between $\mathbf{w}_i$ and $\mathbf{v}_j$, ${\rm T}\in \uppercase\expandafter{\romannumeral2}(\mathbf{a},\mathbf{b}) = \{{\rm T} \in \mathbb{R}^{T \times K} \vert {\rm T}\mathbf{1}_m = \mathbf{a}, {\rm T}^{\top}\mathbf{1}_n = \mathbf{b}\}$, $\mathbf{a}$ and $\mathbf{b}$ Dirac function coefficients centered on $\mathbf{w}_i$ and $\mathbf{v}_j$.

\subsection{Frame Order Modeling}\label{foa}
To better model the timing of the video, VLP models randomly disrupt the order of some input frames and then predict the actual position of each frame. Frame Order Modeling (FOM)~\cite{li2020hero} is modeled as a classification task in practice.

\subsection{Particular Pre-training Objects}\label{vqa}
To better adapt to downstream tasks, VLP models sometimes use the training objects of some downstream tasks, such as visual question answering (VQA)~\cite{antol2015vqa,lei2018tvqa,anderson2018bottom}, and visual captioning (VC)~\cite{vinyals2015show,bai2018survey}, as pre-training objectives. As for VQA, VLP models take the fused representation mentioned above, apply an FC layer, and use the transformed representation to predict the classification over predefined answer candidates. In addition to VLP models tackling the task as classification over predefined answer candidates, VLP models also can directly generate answers in their original text format. As for VC, to reconstruct the input sentence to endow VLP models with the generation capability, VLP models employ an auto-regressive decoder to generate a corresponding textual description of the image or video.


Note that due to space limitations, we only introduce some popular pre-training objectives. We omit some specific pre-training objectives such as grounding referring expression (GRE), image-conditioned denoising autoencoding (IDA)~\cite{xia2021xgpt}, text-conditioned image feature generation (TIFG)~\cite{xia2021xgpt}, object detection (OD)~\cite{kamath2021mdetr} and aligned Kaleido patch modeling (AKPM)~\cite{zhuge2021kaleido}. Moreover, we put masked action prediction into the category of MVM.

\section{Pre-training Datasets}\label{pd}

\begin{table}[tbp]
    \centering
    \caption{Details of some popular pre-training datasets for VLP. Names of some datasets are abbreviated for the convenience of subsequent description. FLKR represents Flickr30k, and HT100M represents HowTo100M.}\label{pre_training_datasets}
    \scalebox{0.90}
    {
        \begin{tabular}{l|ccccc}
            \toprule
            Dataset & \# Images & \# Image-text Pairs & Duration (hrs) & \# Clips & \# Videos \\
            \midrule
            SBU~\cite{ordonez2011im2text} & 875K & 875K & - & - & - \\
            FLKR~\cite{young2014image} & 29K & 145K & - & - & - \\
            COCO~\cite{lin2014microsoft} & 113K & 567K & - & - & -	\\
            VG~\cite{krishna2017visual} & 108K & 5.4M & - & - & - \\
            VGQA~\cite{krishna2017visual} & 108K & 1.8M & -  & - & - \\
            VQA~\cite{goyal2017making} & 83K & 444K & - & - & - \\
            Matterport3D~\cite{chang2017matterport3d} & 104K & 104K & -& - & - \\
            FashionGen~\cite{rostamzadeh2018fashion} & 260K & 260K & - & - & - \\
            CC3M~\cite{sharma2018conceptual} & 3M & 3M & -& - & - \\
            GQA~\cite{hudson2019gqa} & 82K & 1M & -& - & - \\
            LAIT~\cite{qi2020imagebert} & 10M & 10M &	- & - & - \\
            CC12M~\cite{changpinyo2021conceptual} & 12M & 12M & - & - & - \\
            ALIGN~\cite{jia2021scaling} & 1.8B & 1.8B & -& - & - \\
            \midrule
            Kinetics400~\cite{kay2017kinetics} & -  & - & 817 & 306K & 306K \\
            TVQA~\cite{lei2018tvqa} & -  & -  & 461 & 22K & 925 \\
            HT100M~\cite{miech2019howto100m} & - & - & 134K & 136M & 1.2M \\
            WebVid2M~\cite{bain2021frozen} & - & - & 13K & 2.5M & 2.5M \\
            \bottomrule
        \end{tabular}
    }
\end{table}

Pre-training datasets are significant for the success of cross-modal representation learning. The quality and the size of pre-training datasets sometimes overwhelm the importance of training strategies and algorithms. Hence, a detailed description of several widely used pre-training datasets is necessary. Table \ref{pre_training_datasets} shows statistics of some popular pre-training datasets for VLP. 

Since VLP includes image-language pre-training and video-language pre-training, we roughly divide pre-training datasets into two main categories. In later sections, we provide more details about representative pre-training datasets for each category. It is worth noting that no matter which category pre-training datasets belong, they differ in size and sources across different researches. In most works, the pre-training datasets for VLP are constructed by combining public datasets across different cross-modal tasks or scenarios. However, other works, such as VideoBERT \cite{sun2019videobert}, ImageBERT \cite{qi2020imagebert}, ALIGN \cite{jia2021scaling}, and CLIP \cite{radford2021learning}, conduct pre-training with self-constructed datasets. These self-constructed datasets are usually larger than most public datasets but might contain more noise.

\subsection{Datasets for Image-language Pre-training}\label{DILP} 

For image-language pre-training, the most widely used data form is image-text pairs. Most image-language pre-training datasets consist of a large number of image-caption pairs. SBU \cite{ordonez2011im2text} and Flickr30k \cite{young2014image} are collected from Flickr and labelled with human-generated annotations. COCO \cite{lin2014microsoft} consists of images with five human-generated captions, filtered with special procedures to guarantee the quality of images and annotations. CC3M \cite{sharma2018conceptual} and CC12M \cite{changpinyo2021conceptual} are constructed by crawling images and their alt-text HTML attributes from the Internet and annotating these pictures with filtered descriptions. Due to looser filtering strategies, CC12M contains more noise than CC3M. Another data source is the visual question answering task. Many image-language datasets are organized as structured data in the context of visual question answering. The representative large-scale dataset is Visual Genome (VG) \cite{krishna2017visual}. VG contains rich information in its structured data form. Its region-level descriptions and question-answer pairs are frequently used in the study of image-language pre-training. Besides VG, VQA \cite{goyal2017making} and GQA \cite{hudson2019gqa} are also popular datasets of visual question-answer pairs. Compared with VGA, GQA further alleviates the systematic biases. 

Datasets mentioned above are suitable for most common scenarios. There are also some datasets designed for special cases. Matterport3D \cite{chang2017matterport3d} consists of RGB-D images of building-scale scenes, annotated with labels for classification and segmentation. Fashion-Gen \cite{rostamzadeh2018fashion} contains fashion images paired with item descriptions generated by professional stylists.

\subsection{Datasets for Video-language Pre-training}\label{DVLP}

Compared to image-language pre-training datasets, video-language pre-training datasets are usually more time-consuming and more difficult to collect and process. These inconveniences restrict the development of the community and the scale of pre-training. Datasets for video-language pre-training cover different scenarios and sources. Most of them, such as Kinetics-400 \cite{kay2017kinetics}, HowTo100M \cite{miech2019howto100m} and WebVid-2M \cite{bain2021frozen}, are collected from the Internet and processed with different procedures. These kinds of videos are usually accompanied by subtitles, thus providing weak or strong alignments between video clips and text. Although those subtitles sometimes might be too weak to align modalities, they still provide useful information, especially for the pre-training on large-scale datasets. Another source of video-text pairs is television programs. TVQA \cite{lei2018tvqa} is a video-language pre-training dataset generated from television shows. These television shows are collected and converted to a dataset comprised of many dialogues for understanding the videos and recognizing semantic concepts in videos. 

Considering the diversity of the sources and formation of these datasets, researchers apply different annotation and processing procedures. For example, Kinetics-400 \cite{kay2017kinetics} consists of many action-related videos annotated with action classes. For other datasets \cite{lei2018tvqa,miech2019howto100m,bain2021frozen}, the accompanying captions/subtitles of video clips or the class of concepts in videos are usually processed and used as annotations.

\section{Downstream Tasks} \label{dt}
\begin{figure}[t]
\centering
\scalebox{0.99}{
  \begin{overpic}[width=\textwidth]{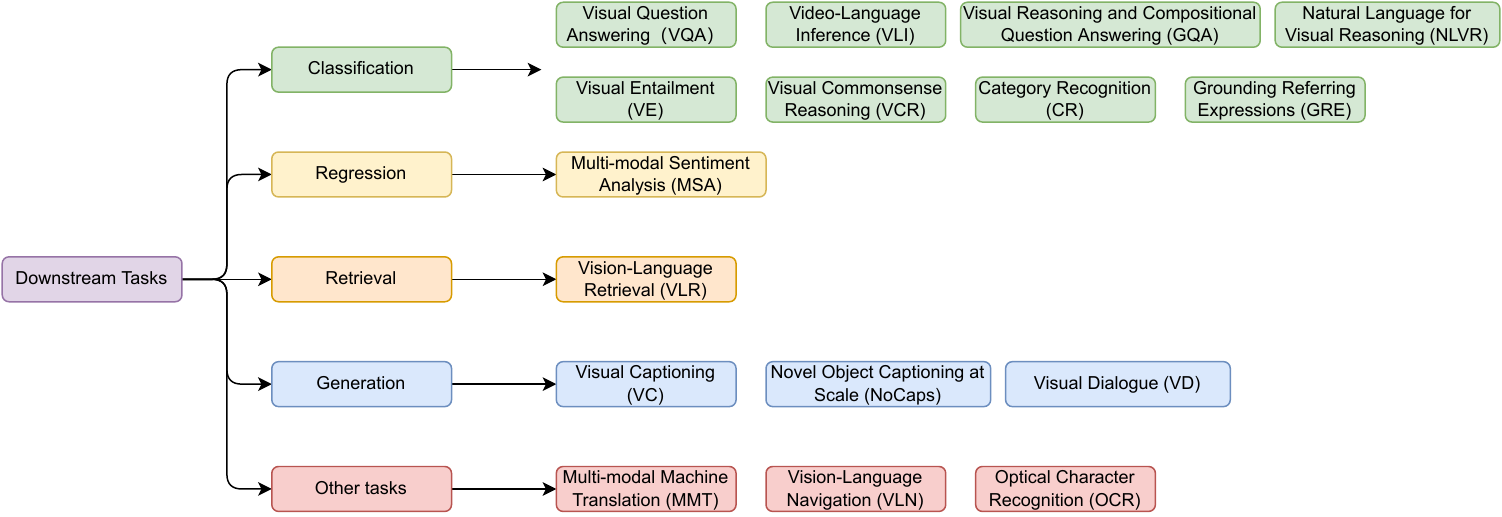}
  \end{overpic}
  }
  \caption{Illustration of downstream tasks in VLP.
  }\label{fig:downstreamtask} 
\end{figure}

As shown in Figure~\ref{fig:downstreamtask}, a diverse range of tasks requires a cooperative knowledge of vision and language. In this section, we introduce the fundamental details and goals of these tasks.

\paragraph{Visual Question Answering (VQA)~\cite{antol2015vqa,wu2017visual,kafle2017visual,kafle2017analysis}}. Giving a visual input (image or video), VQA represents the task of correctly providing an answer to a question. It is usually regarded as a classification task where the model predicts the most suitable answer from a pool of choices. To obtain accurate performance, it is important to infer logical entailments from images (or videos) based on the question posed.


\paragraph{Visual Reasoning and Compositional Question Answering (GQA)~\cite{hudson2019gqa,geng20192nd,bitton2021automatic}}. GQA is an upgraded version of VQA and aims to advance research on the visual reasoning of natural scenes. The images, questions, and answers in its dataset have matching semantic representations. The advantage of this structured representation is that the distribution of answers can be more uniform, and we can analyze the model's performance from more dimensions. Compared with the single evaluation metric (e.g., accuracy) of traditional VQA, GQA includes multi-dimensional evaluation metrics: consistency, validity, plausibility, distribution, and grounding.

\paragraph{Video-Language Inference (VLI)~\cite{li2020hero,li2021adaptive,chaudhary2021robust}}. Given a video clip with aligned subtitles as a premise, paired with a natural language hypothesis based on the video content, a model needs to infer whether the hypothesis is entailed or contradicted by the given video clip.

\paragraph{Visual Entailment (VE)~\cite{xie2019visual,song2022clip,xie2018visual}}. In the VE task, image is the premise, and text is the hypothesis. Its goal is to predict whether the text is ``Entailment Image''. There are three labels, Entailment, Neutral, and Contradiction.

\paragraph{Visual Commonsense Reasoning (VCR)~\cite{zellers2019recognition,yu2019heterogeneous,ye2021case}}. VCR is the task of inferring commonsense information and cognitive understanding by a machine when it sees an image. It exists in the form of multiple-choice questions. For a question posed about the image, there are several alternative answers. The model must choose an answer from several answers and then select the reason for choosing this answer from several alternative reasons. Thus, VCR can be divided into two tasks, including question answering (selecting the best answer from a pool of expected answers to the question) and answer justification (providing the rationale behind the given answer). You can follow VCR's leaderboard\footnote{\url{https://visualcommonsense.com/leaderboard/}} to track VLP's latest ideas.

\paragraph{Natural Language for Visual Reasoning (NLVR)~\cite{suhr2017corpus,marasovic2020natural}}. NLVR is a subtask of the broader VCR category, limited to the classification paradigm. The input of the NLVR task is two images and a text description, and the output is whether the corresponding relationship between the images and the text description is consistent (two labels: true or false). It is typically different from VQA due to longer text sequences covering various linguistic phenomena.

\paragraph{Grounding Referring Expressions (GRE)~\cite{liu2019improving,yang2019cross,zhang2018grounding}}. The GRE task aims to localize certain regions (e.g., objects and persons) in an image given a referring expression, where the main challenge is to comprehend and align various types of information from visual and textual domain, such as visual attributes, locations and interactions with surrounding regions. Specifically, the model can output a score for each region, and the region with the highest score is used as the prediction region.

\paragraph{Category Recognition (CR)~\cite{zhuge2021kaleido}.} CR refers to identifying the category and sub-category of a product, such as \{HOODIES, SWEATERS\}, \{TROUSERS, PANTS\}, which are vital attributes for describing a product, and are useful in lots of real-life applications.

\paragraph{Multi-modal Sentiment Analysis.} (MSA)~\cite{ghosal2018contextual,akhtar2019multi,jiming2021summary,zhang2020knowledge}. MSA is aimed to detect sentiments in videos by leveraging multi-modal signals (e.g., vision, language, etc.). It is to predict the affective orientation of an utterance as a continuous intensity variable.

\paragraph{Vision-Language Retrieval (VLR)~\cite{wang2016comprehensive,mithun2018learning,chen2020imram,chen2022hivlp}.} VLR involves understanding both vision (image or video) and language domains with appropriate matching strategies. It includes two subtasks, vision-to-text, and text-to-vision retrieval, where vision-to-text retrieval is to fetch the top-most relevant text description from a larger pool of descriptions as per the vision and vice versa. VLR is widely used in domain-specific searches, multiple search engines, and context-based vision retrieval design systems. 

\paragraph{Visual Captioning (VC)~\cite{xu2015show,bai2018survey,wang2018reconstruction}.} VC aims to generate semantically and syntactically appropriate text descriptions for a given visual (image or video) input. Generating relevant and explanatory captions for a visual input requires not only a rich knowledge of language, but also a consistent understanding of scenes, entities, and their interactions appreare in the visual input.

\paragraph{Novel Object Captioning at Scale (NoCaps)~\cite{agrawal2019nocaps,feng2020cascaded}}. NoCaps extends the VC task to test a model’s capability of describing novel objects from the Open Images dataset, which are unseen in the training corpus.

\paragraph{Visual Dialogue (VD)~\cite{das2017visual,chen2020dmrm,chen2021gog,chen2021multimodal}.}
The specific task in VD is the
following: given an image, a dialog history consisting of a sequence of question-answer pairs, and
a natural language follow-up question, the goal for the task is to response the question in free-form natural language (e.g., generate an answer). VD is the visual analogue of the Turing Test. 

\paragraph{Multi-modal Machine Translation (MMT)~\cite{specia2016shared,yin2020novel,su2019unsupervised}.} MMT is a two-fold task of translation and text generation, translating text from one language to another with additional information from other modalities, e.g., image. The additional visual features aim to remove ambiguities that may arise in straightforward text machine translation and help retain the context of the text descriptions. The multi-modal representation space facilitates robust latent representations to complement the inherent semantic information preserved by visual and linguistic embeddings, respectively.

\paragraph{Vision-Language Navigation (VLN)~\cite{wang2019reinforced,zhu2020vision,gu2022vision}.} VLN is a grounding language task of an agent’s locomotion as it sees and explores the real-world dynamics based on linguistic instructions. Like generation tasks, it is typically seen as the task of sequence-to-sequence transcoding. However, VLN has unique characteristics. It usually has longer sequences, and the dynamics of the problem are quite different since it is a real-time evolving task. Its main challenge lies in understanding the environment and making confident decisions during exploring.


\paragraph{Optical Character Recognition (OCR)~\cite{mori1999optical,memon2020handwritten}.} OCR generally refers to extract handwritten or printed text from images (such as street signs and photos of products) as well as documents (articles, bills, invoices, financial reports, etc.), which includes two parts: text detection (similar to regression) and text recognition (similar to classification).

In addition, there are some iamge-related downstream tasks for evaluating the image-text pre-training models, including semantic segmentation~\cite{strudel2021segmenter,mo2022review}, and object detection~\cite{zhao2019object,fang2021you}. There are also some video-related downstream tasks for evaluating the video-text pre-training models, including action classification (AC)~\cite{sun2019videobert}, action segmentation (AS)~\cite{sun2019learning}, and action step Localization (ASL)~\cite{luo2020univl}. 

Recently, Changpinyo et.al~\cite{changpinyo2021conceptual} scale up pre-training data for VLP tasks and benchmark its effectiveness against Conceptual Captions 3M on multiple downstream tasks with an emphasis on long-tail visual recognition. Rethmeier et.al~\cite{rethmeier2022long} study the performance of pretrained model on a challenging long-tail task and analyze the resulting long-tail learning capabilities under zero-shot, few-shot and full supervision conditions to explore the performance influence of model size and self-supervision signal amount.

\begin{sidewaystable}
\sidewaystablefn%
\begin{center}
\begin{minipage}{\textheight}
\caption{The summary of mainstream image-text VLP models. The number of downstream tasks determines whether the model is generic or domain-specific VLP. FE: Feature Extraction. PT: Pre-training. Emb: Embedding. SC in Datatsets column: self-constructed or self-collected. MTL in Datatsets column: all datasets for multi-task learning in corresponding work. See other abbreviations in Datatsets column in Table \ref{pre_training_datasets}.}\label{taxonomy}
\resizebox{0.99\textwidth}!{
\begin{tabular}{l|cccccccc}
\toprule
Model & Domain & Vision FE & Language FE & Multimodal Fusion & Decoder & PT Objectives & PT Datasets & Downstream Tasks \\
\midrule
VisualBERT~\cite{li2019visualbert} &	Image&	OD-RFs&	Emb&	Single-stream&	No&	MLM+VLM&	COCO&	GRE+NLVR+VCR+VQA\\
ViLBERT~\cite{lu2019vilbert}  &   Image   &   OD-RFs &   Emb	&   Dual-stream &   No  & MLM+VLM+MVM   & COCO+VG   & VLR+NLVR+VE+VQA\\
LXMERT~\cite{tan2019lxmert}  &   Image &	OD-RFs+Xformer  &   Xformer &	Dual-stream &	No &	MLM+VLM+MVM+VQA	& COCO+VG+VQA+GQA+VGQA & GQA+NLVR+VQA \\
B2T2~\cite{alberti2019fusion}&	Image&	 CNN-GFs&	Emb&	Single-stream&	No&	MLM+VLM&	CC3M&	VCR\\
Unicoder-VL~\cite{li2020unicoder}&	Image&	OD-RFs&	Emb&	Single-stream&	No&	MLM+VLM+MVM&	CC3M+SBU&	VLR+VCR\\
VL-BERT~\cite{su2019vl}&	Image&	OD-RFs&	Emb&	Single-stream&	No&	MLM+MVM&	CC3M&	GRE+VCR+VQA\\
VLP~\cite{zhou2020unified}&	Image&	OD-RFs&	Emb&	Dual-stream&	Yes&	MLM+LM&	CC3M&	VC+VQA\\
UNITER~\cite{chen2020uniter}&	Image&	OD-RFs&	Emb&	Single-stream&	No&	MLM+VLM+MVM+WRA&	COCO+VG+SBU+CC3M&	GRE+VLR+NLVR+VCR+VE+VQA\\
12-IN-1~\cite{lu202012}	&Image&	OD-RFs&	Emb&	Single-stream&	No&	MLM+MVM&	MTL&	GQA+GRE+VC+NLVR+VE+VQA\\
VisDial-BERT~\cite{murahari2020large}&	Image&	OD-RFs&	Emb&	Dual-stream&	No&	MLM+VLM+MVM&	CC3M+VQA&	VD\\
ImageBERT~\cite{qi2020imagebert}&	Image&	OD-RFs&	Emb&	Single-stream&	No&	MLM+VLM+MVM&	LAIT+CC3M+SBU &	VLR\\
PREVALENT~\cite{hao2020towards}&	Image&	 CNN-GFs+Xformer&	Xformer&	Single-stream&	No&	MLM+MVM&	Matterport3D&	VLN\\
XGPT~\cite{xia2021xgpt}&	Image&	OD-RFs&	Emb& Dual-stream&	Yes&	MLM+IDA+VC+TIFG&	CC3M&	VC+VLR\\
InterBER~\cite{lin2020interbert}&	Image&	OD-RFs&	Emb&	Single-stream&	No&	MLM+VLM+MVM&	COCO+CC3M+SBU&	VLR+VCR\\
PixelBERT~\cite{huang2020pixel}&	Image&	 CNN-GFs&	Emb&	Single-stream&	No&	MLM+VLM&	COCO+VG&	VLR+NLVR+VQA\\
OSCAR~\cite{li2020oscar}&	Image&	OD-RFs&	Emb&	Single-stream&	No&	MLM+VLM&	COCO+SBU+CC3M+FLKR+VQA+GQA+VGQA&	GQA+VC+VLR+NLVR+NoCaps+VQA\\
VLN-BERT~\cite{hong2021vln}&	Image&	OD-RFs&	Emb&	Dual-stream&	No&	MLM+VLM+MVM&	CC3M&	VLN\\
FashionBERT~\cite{gao2020fashionbert}&	Image&	Xformer&	Emb&	Single-stream&	No&	MLM+VLM+MVM&	FashionGen&	VLR\\
VILLA~\cite{gan2020large}&	Image&	OD-RFs+Xformer&	Xformer&	Single-stream&	No&	MLM+VLM+MVM&	COCO+VG+CC3M+SBU&	GRE+VLR+NLVR+VCR+VE+VQA\\
ERNIE-ViL~\cite{yu2020ernie}&	Image&	OD-RFs&	Emb&	Single-stream&	No&	MLM+MVM&  	CC3M+SBU&	GRE+VLR+VCR+VQA\\
RVL-BERT~\cite{chiou2021visual}&	Image&	OD-RFs&	Emb&	Single-stream&	No&	MLM+VLM+MVM&	CC3M&	VC+VQA\\
VinVL~\cite{zhang2021vinvl}&	Image&	OD-RFs&	Emb&	Single-stream&	No&	MLM+VLM&	COCO+CC3M+SBU+FLKR+VQA+GQA+VGQA&	GQA+VC+VLR+NLVR+NoCaps+VQA\\
VL-T5~\cite{cho2021vlt5}&	Image&	OD-RFs&	Emb&	Single-stream&	Yes&	MLM+VLM+VQA+GRE+VC&	COCO+VG+VQA+GQA+VGQA&	GQA+GRE+VC+MMT+NLVR+VCR+VQA\\
ViLT~\cite{kim2021vilt}&	Image&	ViT-PFs&	Emb&	Single-stream	&No&	MLM+VLM&	COCO+VG+SBU+CC3M&	VLR+NLVR+VQA\\
ALIGN~\cite{jia2021scaling}&	Image&	 CNN-GFs&	Xformer&	Dual-stream&	No&	VLC&	ALIGN &	VLR\\
Kaleido-BERT~\cite{zhuge2021kaleido}&	Image&	 CNN-GFs&	Emb&	Single-stream&	No&	MLM+VLM+AKPM&	FashionGen&	CR+VC+VLR\\
MDETR~\cite{kamath2021mdetr}&	Image&	Xformer&	Xformer&	Single-stream&	Yes&	OD+MLM+VLC&	COCO+VG+FLKR+GQA&	GQA+VQA\\
SOHO~\cite{huang2021seeing}&	Image&	 CNN-GFs&	Emb&	Single-stream&	No&	MLM+VLM+MVM&	COCO+VG&	VLR+NLVR+VE+VQA\\
E2E-VLP~\cite{xu2021e2e}&	Image&	 CNN-GFs&	Emb&	Single-stream&	Yes&	OD+MLM+VLM&	COCO+VG&	VC+VLR+NLVR+VQA\\
Visual Parsing~\cite{xue2021probing}&	Image&	Xformer&	Emb&	Single-stream&	No&	MLM+VLM+MVM&	COCO+VG&	VLR+VCR+VE+VQA\\
CLIP-ViL~\cite{shen2021much}&	Image&	 CNN-GFs&	Emb&	Single-stream&	Yes&	MLM+VLM+VQA&	COCO+VG+VQA+GQA+VGQA&	VE+VLN+VQA\\
ALBEF~\cite{li2021align}&	Image&	Xformer&	Xformer&	Dual-stream&	No&	MLM+VLM+VLC&	COCO+VG+CC3M+SBU&	VLR+NLVR+VQA\\
SimVLM~\cite{wang2021simvlm}&	Image&	 CNN-GFs&	Emb&	Single-stream&	Yes&	PrefixLM&	ALIGN&	VC+NLVR+VE+VQA\\
MURAL~\cite{jain2021mural}&	Image&	 CNN-GFs&	Xformer&	Dual-stream&	No&	VLC&	CC12M+ALIGN&	VC+VLR\\
VLMO~\cite{vlmo} &	Image &	ViT-PFs &	Emb&	Single-stream&	No &	MLM+VLC+VLM&	COCO+VG+CC3M+SBU&	VQA+NLVR+VLR\\
METER~\cite{dou2021empirical}&	Image&	Xformer&	Xformer&	Dual-stream&	No&	MLM+VLM&	COCO+VG+CC3M+SBU&	VLR+NLVR+VE+VQA\\
X-VLM~\cite{zeng2021multi}&	Image&	Xformer&	Xformer&	Single-stream&	No&	MLM+VLM+VG&	COCO+VG+CC3M+SBU&	VLR+NLVR+VE+VQA\\
TCL~\cite{yang2022vision}&	Image&	Xformer&	Xformer&	Single-stream&	No&	MLM+VLM+TCL&	COCO+VG+CC3M+SBU&	VLR+NLVR+VE+VQA\\
\bottomrule
\end{tabular}
}
\end{minipage}
\end{center}
\end{sidewaystable}

\begin{sidewaystable}
\sidewaystablefn%
\begin{center}
\begin{minipage}{\textheight}
\caption{The summary of mainstream video-text VLP models. The number of downstream tasks determines whether the model is generic or domain-specific VLP. FE: Feature Extraction. PT: Pre-training. Emb: Embedding. SC in Datatsets column: self-constructed or self-collected. MTL in Datatsets column: all datasets for multi-task learning in corresponding work. See other abbreviations in Datatsets column in Table \ref{pre_training_datasets}.}\label{taxonomy2}
\resizebox{0.99\textwidth}!{
\begin{tabular}{l|cccccccc}
\toprule
Model & Domain & Vision FE & Language FE & Multimodal Fusion & Decoder & PT Objectives & PT Datasets & Downstream Tasks \\
\midrule
VideoBERT~\cite{sun2019videobert}  & Video  & CNN-GFs  & Emb  &	Single-stream	 &No  & MLM+VLM+MVM & 	SC & AC+VC\\
CBT~\cite{sun2019learning}&	Video&	 CNN-GFs+Xformer&	Xformer&	Single-stream&	No&	VLC&	Kinetics&	AC+AS+VC\\
UniVL~\cite{luo2020univl}&	Video&	 CNN-GFs& Xformer&	Dual-stream&	Yes&	MLM+VLM+VC&	HT100M &	AS+ASL+MSA+VC+VLR\\
HERO~\cite{li2020hero}&	Video&	 CNN-GFs+Xformer&	Xformer&	Single-stream&	No&	MLM+VLM+MVM+FOM&	HT100M+TV&	VC+VLI+VQA+VLR\\
MMFT-BERT~\cite{urooj2020mmft}&	Video&	OD-RFs+Xformer&	Xformer&	Single-stream&	No&	VQA &	TV&	VQA\\
ActBERT~\cite{zhu2020actbert}&	Video&	OD-RFs+CNN&	Emb&	Single-stream&	No&	MLM+VLM+MVM&	HT100M&	AS+ASL+VC+VQA+VLR\\
CLIP~\cite{radford2021learning}&	Image / Video&	CNN/Xformer&	Xformer&	Dual-stream&	No&	VLC&	SC& OCR +AC etc.\\
Frozen~\cite{bain2021frozen} & Video & ViT-PFs & Emb & Dual-Stream & No & VLC & WebVid2M+CC3M& VLR\\
Region-Learner~\cite{yan2021video}& Video & ViT-PFs & Emb & Dual-Stream & No & VLC & WebVid2M+CC3M& VLR\\
CLIP4Clip~\cite{luo2021clip4clip}& Video & ViT-PFs & Emb & Dual-Stream & No & VLC & WebVid2M+CC3M& VLR\\
CLIP2Video~\cite{fang2021clip2video}& Video & ViT-PFs & Emb & Dual-Stream & No & VLC & WebVid2M+CC3M& VLR\\
\bottomrule
\end{tabular}
}
\end{minipage}
\end{center}
\end{sidewaystable}

\section{SOTA VLP models} \label{vlp}

\paragraph{Image-Text VLP models.}
VisualBERT~\cite{li2019visualbert}, known as the first image-text pre-training model, uses the visual features extracted by Faster R-CNN, concatenates the visual features and textual embeddings, and then fed the concatenated features to a single transformer initialed by BERT. Many VLP models~\cite{li2020unicoder,su2019vl,chen2020uniter,qi2020imagebert} follow the similar feature extraction and architecture as VisualBERT while adjusting the pre-training objectives and pre-training datasets. Recently, VDBERT~\cite{wang2020vd} models the common implicit vision-language alignment in vision and language by pretraining on large-scale image-text pairs via transfer learning~\cite{dong2020can,dong2021and}. VLMO~\cite{vlmo} leverages patch embeddings for image and word embeddings for text and feeds the concatenated embeddings into a single transformer with modality experts and achieves an impressive performance. METER \cite{dou2021empirical} explores how to use a uni-modal pre-trained model and proposes a dual-stream architecture model to handle the multimodel fusion, which achieves the SOTA performance on many downstream tasks. The summary of mainstream image-text VLP models is shown in Table~\ref{taxonomy}.

\paragraph{Video-Text VLP models.} VideoBERT~\cite{sun2019videobert}, known as the first video-text pre-training model, extends the BERT model to process videos and texts simultaneously. VideoBERT uses the pre-trained ConvNet and S3D~\cite{xie2017rethinking} to extract video features and concatenate them with textual word embeddings to feed into a transformer initialed with BERT. ConvNet and S3D are frozen when training the VideoBERT, which indicates the approach is not end-to-end. Recently, inspired by ViT, CLIP4Clip~\cite{luo2021clip4clip} and CLIP2Video~\cite{fang2021clip2video} first process video clips into frames and get patch embeddings according to the method of ViT processing images for each frame. CLIP4clip and CLIP2Video optimize themselves in an end-to-end manner and achieve SOTA performance. The summary of mainstream video-text VLP models is shown in Table~\ref{taxonomy2}.



\section{Conclusion and New Frontiers} \label{nf}
In this paper, we provide the first VLP survey. We review its recent advances from five aspects: feature extraction, model architecture, pre-training objectives, pre-training datasets, and downstream tasks and summarize the specific SOTA VLP models in detail. We hope our survey can help researchers understand VLP better and inspire new works to advance this field. In the future, based on existing works, VLP can be further developed from the following aspects: 

\paragraph{Incorporating Acoustic Information.} Most previous works on multi-modal pre-training emphasize the joint modeling of language and vision but ignore the information buried in audios~\cite{zhu2021deep,tao2021correction}. Although the semantic information in audios might intersect with language, audios could provide extra emotion information, acoustic boundary information, etc. Moreover, pre-training with audios makes the model capable of downstream tasks with acoustic inputs. Until now, joint modeling and representation across text, vision, and audio is still an open problem left for further investigation. Several cutting-edge works have shed light on the future of this research field. Unlike previous VLP models, VATT \cite{akbari2021vatt} takes the raw audio as input and learns the multi-modal representations with the noise contrastive estimation (NCE). Differing from VATT, OPT \cite{liu2021opt} learns the cross-modal representations across text, image, and audio jointly with various multi-level masking strategies, and it is also capable of generating text and images. Some other works, such as AudioCLIP \cite{guzhov2021audioclip} and MERLOT Reserve \cite{zellers2022merlot}, also shows their unique approaches to learn the cross-modal representations over three modalities.

\paragraph{Knowledgeable and Cognitive Learning.} Although the existing VLP models have achieved remarkable performance, their essence is to fit large-scale multimodal datasets. Making VLP models more knowledgeable is important for future VLP. For input vision and text, there is rich related external common sense world knowledge and illustrative situational knowledge~\cite{chen2021kbvlp}, which can be used to augment the input and accelerate the model training and inference. The solution to this problem requires unified cognitive model architectures, knowledge-guided pre-training objectives, and the support of interacting with new knowledge.
 
\paragraph{Prompt Tuning.} Currently, fine-tuning is the dominant method to transfer the knowledge of VLP to downstream tasks. However, as the scale of the model increases, each downstream task has its fine-tuning parameters leading to parameter inefficiency. Moreover, the diverse downstream tasks also make the design of the pre-training and fine-tuning stages cumbersome, leading to a gap between them. Recently, prompt tuning is getting more and more attention in NLP. By designing discrete or continuous prompts and using MLM for specific downstream tasks, these models could: 1) reduce the computational cost on fine-tuning the enormous amounts of parameters; 2) bridge the gap between pre-training and fine-tuning. Prompt tuning is a promising way to stimulate the linguistic and world knowledge distributed in PLMs. In the next step, it can be improved and transferred to multi-modal scenarios, breaking the traditional paradigm and solving the pain points of VLP \cite{tsimpoukelli2021multimodal}.  

\paragraph{Model Compression and Acceleration.} Model compression and acceleration is an essential approach to improve the efficiency of VLP models. In this case, large models are compressed to small ones to meet the need for faster inference and deployment on various real-life scenarios such as resource-constrained devices. In general PLMs, model compression and acceleration is a hot topic, and specific methods include parameter sharing~\cite{lan2019albert}, model pruning~\cite{fan2019reducing}, knowledge distillation~\cite{sanh2019distilbert} and model quantization~\cite{zafrir2019q8bert}. Recently, knowledge distillation has been used to compress VLP models~\cite{fang2021compressing}, but other methods such as pruning and quantization of VLP models remain to be explored. Furthermore, a data-efficient VLP paradigm is constructed~\cite{li2021supervision}. However, only a few efforts are currently focused on improving the efficiency of VLP models, leaving much room for exploration.

\paragraph{Out-of-domain Pretraining.} Despite the significant progress achieved by VLP models, part of their success can be traced back to the introduction of in-domain pretraining datasets, used in both pretraining and downstream tasks. The out-of-domain pretraining will be an essential research direction, that is, VLP models transfer the learned knowledge and representation into downstream tasks with unknown data distributions. To mitigate the distribution biases between pretraining and funetuning, DeVLBert~\cite{zhang2020devlbert} is proposed to perform intervention-based learning. It borrows the idea of the backdoor adjustment from the research area of causality and designs several neural-network based structures for Bert-style out-of-domain pretraining.


\paragraph{Advanced Model Architecture.}
Nowadays, the transformer-based architectures make great progress in VLP. Is such a structure the optimal structure for VLP? We note that the recently popular diffusion model~\cite{saharia2022photorealistic} for image generation has succeeded greatly. Some researchers~\cite{li2022diffusion} also extend the diffusion model to controllable text generation. So whether the diffusion model can be used in VLP? It may be a question worth exploring in the future. Moreover, neural networks themselves are inspired by neuroscience, and we can explore next-generation VLP frameworks with support from other disciplines. The inspirations from mathematics include the framework of non-Euclidean space Manifold and how to put some geometric priors into the model~\cite{chen2022fully,bronstein2021geometric}, which are relatively new research directions. Research on the energy-efficient Spiking Neural Networks~\cite{maass1997networks,zhang2022recent,zhang2021population} in the brain-inspired field may also provide insights into the exploration of novel VLP architectures.


\bibliographystyle{bst/sn-mathphys}
\bibliography{sn-bibliography}

\begin{figure}[H]%
\centering
\includegraphics[width=0.3\textwidth]{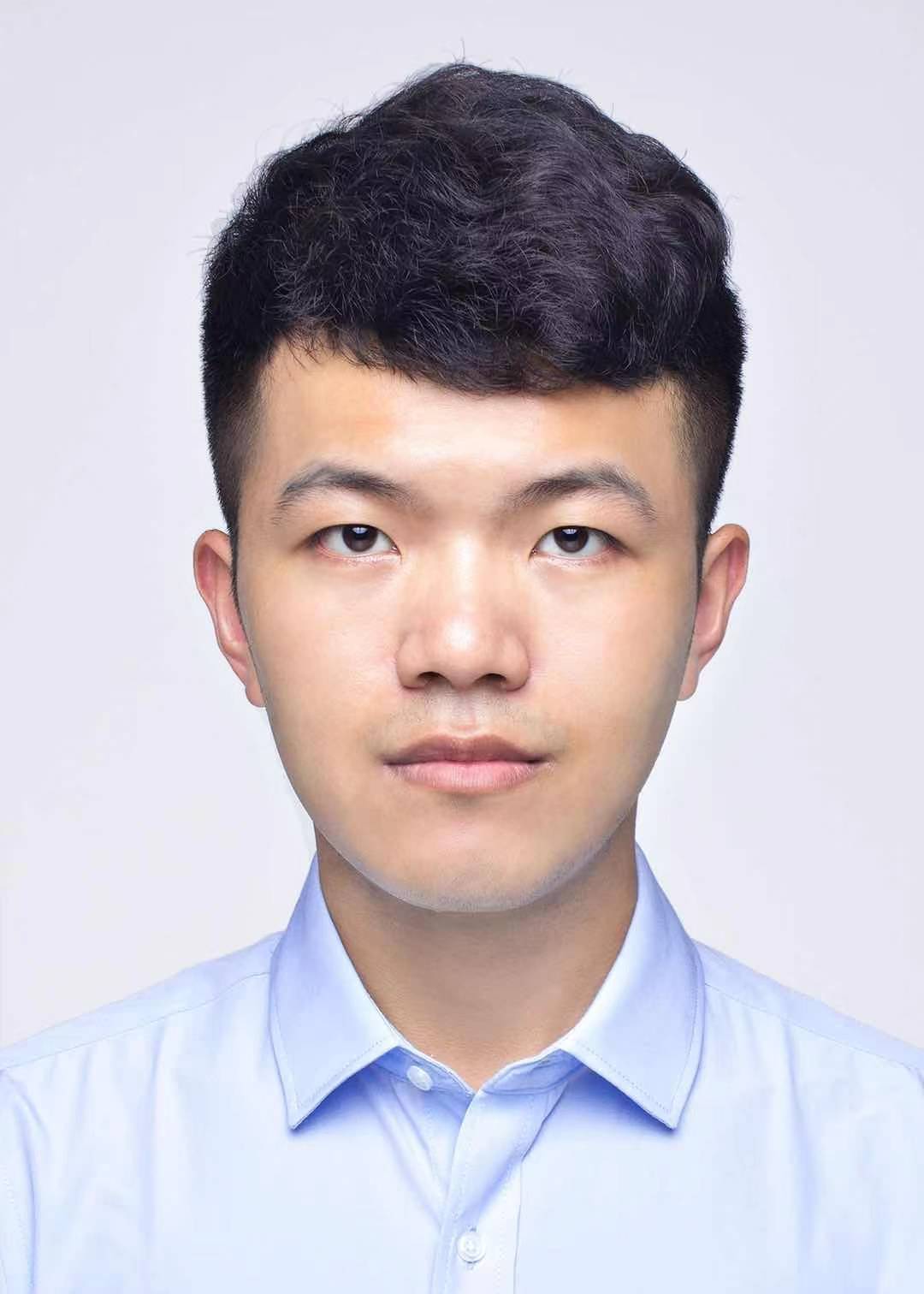}
\end{figure}

\noindent{\bf Feilong Chen}\quad received the B.Sc. degree in computer sciences from Hefei University of Technology, China in 2018. He is a Ph.D. candidate in both the Institute of Automation Chinese Academy of Sciences and the University of Chinese Academy of Sciences. His current interests include theoretical research on vision-language pre-training, multi-modal question answering and dialog.

E-mail: chenfeilong2018@ia.ac.cn

\begin{figure}[H]%
\centering
\includegraphics[width=0.3\textwidth]{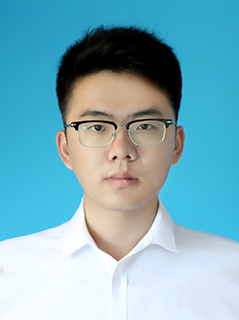}
\end{figure}

\noindent{\bf Duzhen Zhang}\quad received the B.Sc. degree in software engineering from Shandong University, China in 2019. He is a Ph.D. candidate in both the Institute of Automation Chinese Academy of Sciences and the University of Chinese Academy of Sciences. His current interests include theoretical research on reinforcement learning, natural language processing, and Spiking Neural Networks.

E-mail: zhangduzhen2019@ia.ac.cn

\begin{figure}[H]%
\centering
\includegraphics[width=0.3\textwidth]{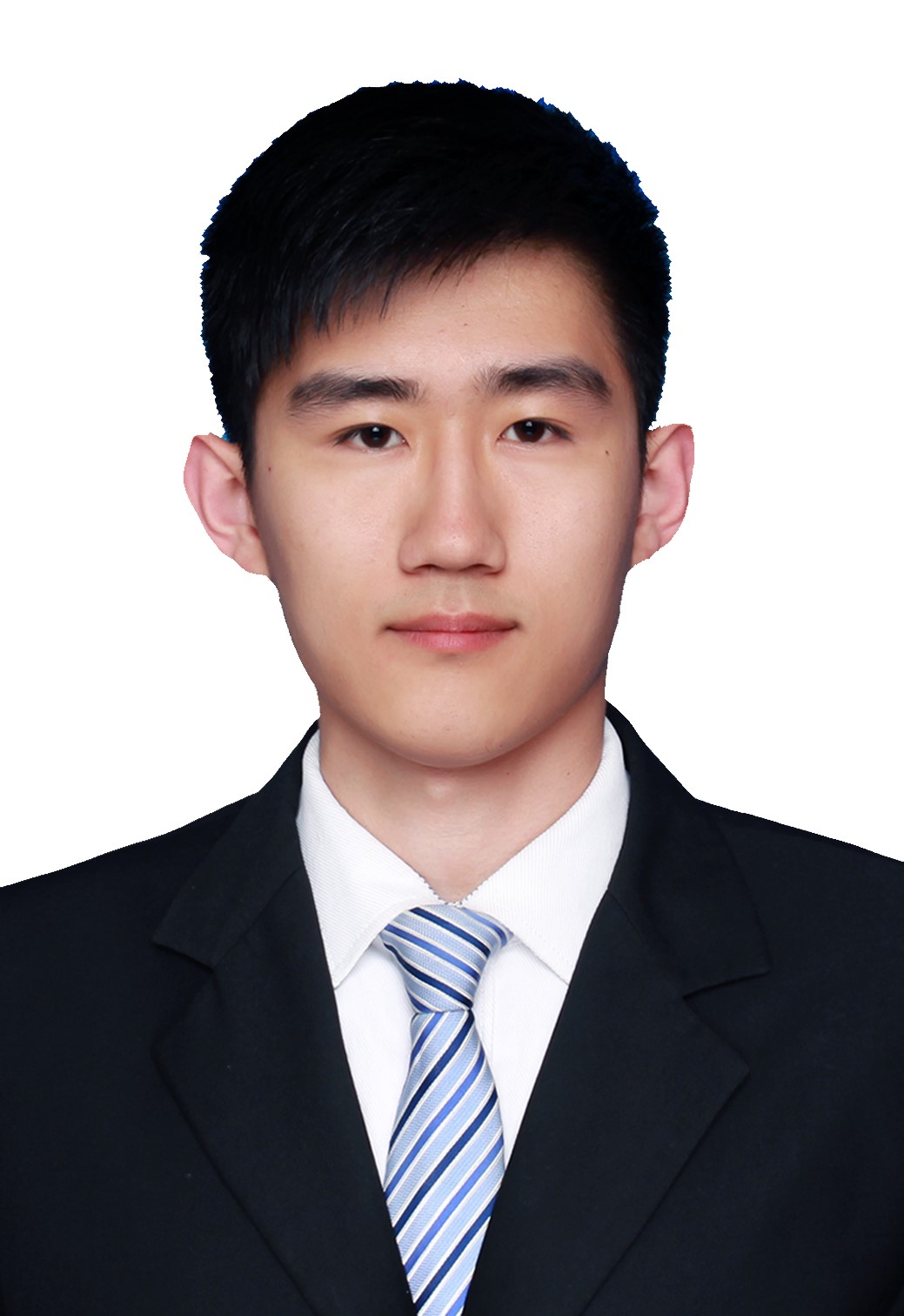}
\end{figure}

\noindent{\bf Minglun Han}\quad received the B.Sc. degree in electronic and information engineering from Harbin Institue of Technology at Weihai, China in 2018. He is a Ph.D. candidate in both the Institute of Automation, Chinese Academy of Sciences and the University of Chinese Academy of Sciences. His current research interests include speech recognition, speech synthesis, speech chain.

E-mail: hanminglun2018@ia.ac.cn

\begin{figure}[H]%
\centering
\includegraphics[width=0.3\textwidth]{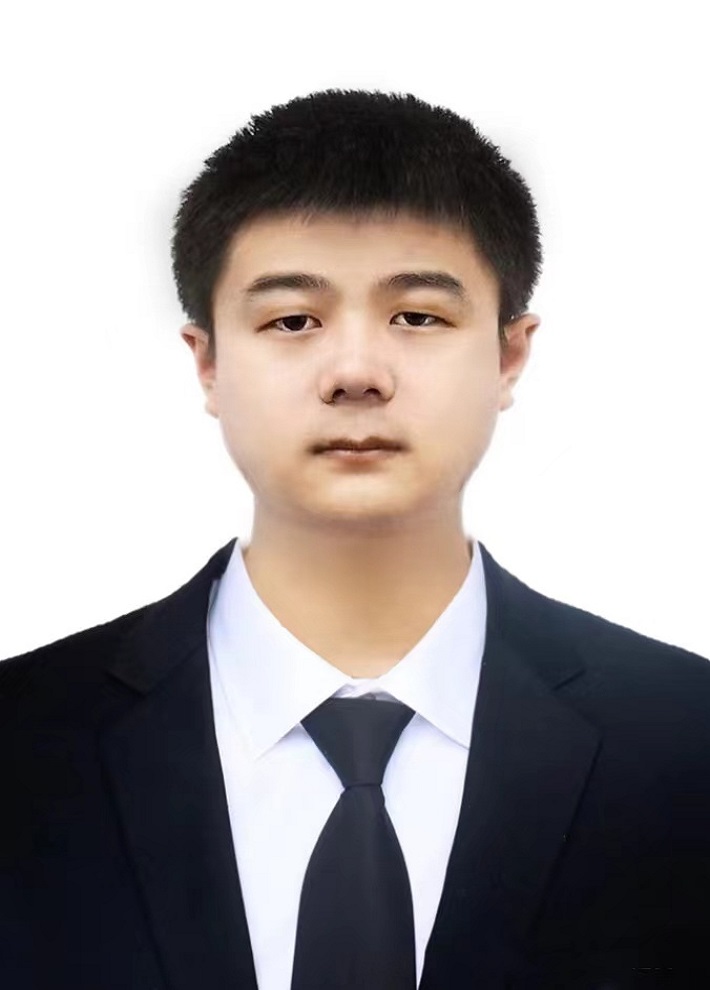}
\end{figure}

\noindent{\bf Xiuyi Chen}\quad received his Ph.D. degree (2022) in Pattern Recognition and Intelligent System from Institute of Automation, Chinese Academy of Sciences, advised by Prof. Bo Xu.
Previously, he received the B.Sc. degree (2017) in Department of Control Science and Engineering from JiLin University. 
His current interests include Cross-modal Retrieval, Multimodal Learning, Dialogue System, Knowledge-Grounded Generation and Speech Seperation.

E-mail: chenxiuyi2017@ia.ac.cn


\begin{figure}[H]%
\centering
\includegraphics[width=0.3\textwidth]{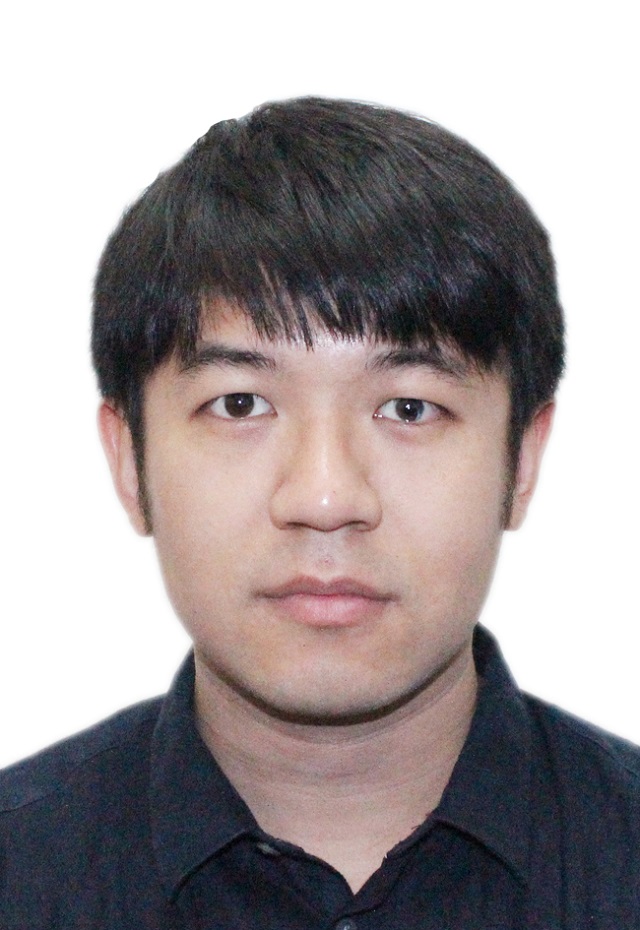}
\end{figure}

\noindent{\bf Jing Shi}\quad is a research assistant in the Institute of Automation, Chinese Academy of Sciences, where he received his Ph.D. degree (2021) in the major of Pattern Recognition and Intelligent System, advised by Prof. Bo Xu.
Previously, he received the B.Sc. degree (2012) in School of Instrumentation and Optoelectronic Engineering from Beihang University. 
His current interests include Cross-modal Modeling, Multimodal Learning, Dialogue System, Speech Recognition and Speech Seperation.

E-mail: shijing2014@ia.ac.cn


\begin{figure}[H]%
\centering
\includegraphics[width=0.3\textwidth]{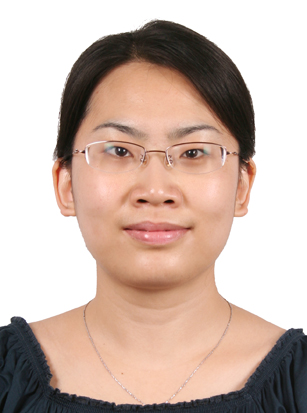}
\end{figure}

\noindent{\bf Shuang Xu}\quad is a professor in Institute of Automation, Chinese Academy of Science. Her main research interests include natural language processing and understanding, human-AI hybird intelligence.

E-mail: shuang.xu@ia.ac.cn

\begin{figure}[H]%
\centering
\includegraphics[width=0.3\textwidth]{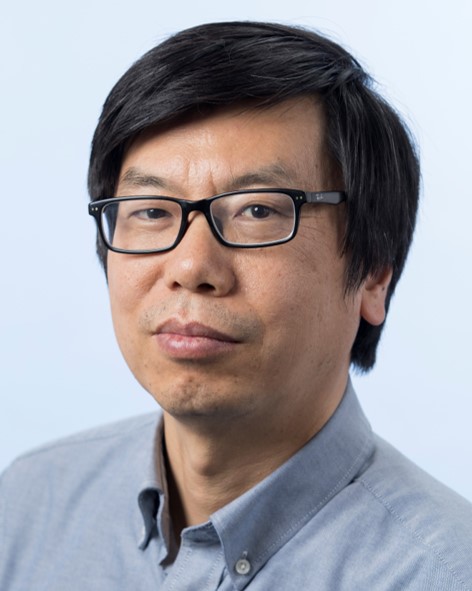}
\end{figure}

\noindent{\bf Bo Xu}\quad is a professor, the director of the Institute of Automation Chinese Academy of Sciences, and also deputy director of the Center for Excellence in Brain Science and Intelligence Technology, Chinese Academy of Sciences. His main research interests include brain-inspired intelligence, brain-inspired cognitive models, natural language processing and understanding, brain-inspired robotics.

E-mail: xubo@ia.ac.cn (Corresponding author)


\end{document}